\documentclass{article}

\usepackage[utf8]{inputenc} % allow utf-8 input
\usepackage[T1]{fontenc}    % use 8-bit T1 fonts
\usepackage{hyperref}       % hyperlinks
\usepackage{url}            % simple URL typesetting
\usepackage{booktabs}       % professional-quality tables
\usepackage{amsfonts}       % blackboard math symbols
\usepackage{nicefrac}       % compact symbols for 1/2, etc.
\usepackage{microtype}      % microtypography
\usepackage{verbatim} 
\usepackage{float}
\usepackage{textcomp}
\usepackage{mathtools}
\usepackage{amsmath}
\usepackage{amssymb}
\usepackage{graphicx}
\usepackage{verbatim}
\usepackage{algorithm} 
\usepackage{bm}
\usepackage{caption}
\usepackage{subfigure}

\usepackage[british]{babel}
\usepackage{bbm}\usepackage{xcolor}\usepackage{cite}
\usepackage{color}\usepackage{amsfonts}\usepackage{algorithm}
\usepackage{algpseudocode}\usepackage{graphicx}
\usepackage{textcomp}
\usepackage{babel}

\newtheorem{proposition}{Proposition}

\newcommand{\ba}{\mathbf{a}}
\newcommand{\bB}{\mathbf{B}}

\newcommand{\bc}{\mathbf{c}}
\newcommand{\bd}{\mathbf{d}}

\newcommand{\br}{\mathbf{r}}
\newcommand{\bR}{\mathbf{R}}

\newcommand{\bX}{\mathbf{X}}
\newcommand{\bx}{\mathbf{x}}

\newcommand{\by}{\mathbf{y}}

\newcommand{\GP}{\text{GP}}

\newcommand{\cA}{\mathcal{A}}

\newcommand{\cX}{\mathcal{X}}

\newcommand{\cR}{\mathcal{R}}

\newcommand{\bbR}{\mathbb{R}}
\newcommand{\bbE}{\mathbb{E}}
\newcommand{\bbI}{\mathbb{I}}

\newcommand{\bdelta}{\boldsymbol{\delta}}

\newcommand{\btheta}{\boldsymbol{\theta}}

\newcommand{\bphi}{\boldsymbol{\phi}}

\newcommand{\Cov}{\mathrm{Cov}}

\newcommand{\argmax}{\operatornamewithlimits{argmax}}

\newcommand{\argmin}{\operatornamewithlimits{argmin}}

\title{Multi-level Training and Bayesian Optimization for Economical Hyperparameter Optimization}

\author{Yang Yang \thanks{yyang15@mails.tsinghua.edu.cn, supported in part by China Scholarship Council (No. 201806210168).} \\
 Department of Mathematical Sciences, Tsinghua University\\
 \and
  Ke Deng \thanks{kdeng@tsinghua.edu.cn,  supported in part by National Natural Science Foundation of China (No 11931001 \& 11771242, DK PI).}\\
  Center for Statistical Sciences, Tsinghua University \\   
  \and 
  Michael Zhu \thanks{yuzhu@stat.purdue.edu}\\
  Department of Statistics, Purdue University  \\
}

\date{}

\begin{document}

\maketitle

\begin{abstract}
	Hyperparameters play a critical role in the performances of many machine learning methods. Determining their best settings or Hyperparameter Optimization (HPO)
	faces difficulties presented by the large number of hyperparameters as well as the excessive training time. In this paper, we develop an effective approach to reducing the total amount of required training time for HPO. In the initialization, the nested Latin hypercube design is used to select hyperparameter configurations for two types of training, which are, respectively, heavy training and light training. We propose a truncated additive Gaussian process model to calibrate approximate performance measurements generated by light training, using accurate performance measurements generated by heavy training. Based on the model, a sequential model-based algorithm is developed to generate the performance profile of the configuration space as well as find optimal ones. Our proposed approach demonstrates competitive performance when applied to 
	optimize synthetic examples, support vector machines, fully connected networks and convolutional neural networks.
\end{abstract}


\section{Introduction}
\label{sec:intro}

Machine Learning (ML) has increasingly become a major force behind modern data-driven technologies ranging from intelligent mobile devices, online marketing, auto-driving to robotics. In particular, Deep Neural Networks (DNNs) have become the choice of prototype models for Deep Learning (DL), thanks to their state-of-the-art performances in complex computer vision and natural language processing tasks. The performances of many machine learning algorithms rely on certain hyperparameters. For example, the training loss of neural networks depends on the hyperparameters including number of hidden layers, number of units per layer, dropout rates, etc.

% Definition of HPO
Considering a machine learning algorithm $\cA$ with $d$ hyperparameters, we denote the 
\emph{hyperparameter configuration space} as $\cX\in \bbR^d$ and a hyperparameter configuration as $\bx\in\cX$. The performance measurement (i.e., the validation loss) of $\cA$ under configuration $\bx$ is denoted by $y(\bx)$, which is the objective function of interest. As $\bx$ varies, $y(\bx)$ gives the performance profile of $\cX$. In practice, $y(\cdot)$ is unknown and needs to be learned from simulation or experiment. For a given ML task, even when a prototype model (e.g. DNN) has already been chosen, training and determining the final model that best suits the ML task at hand can be challenging. One needs to consider and train a large number of configurations of the prototype model, and then select the best model. The \emph{hyperparameter optimization} (HPO) in the ML literature can be formulated as $\bx^* = \text{argmin}_{\bx \in \cX} y(\bx)$.

% Chanllenges of HPO and related literatures
There are two primary difficulties HPO faces. The first difficulty is the curse of dimensionality caused by the different types of hyperparameters and their total numbers. 
%The space of possible configurations of a prototype model such as DNNs not only can be of high dimensions, it can also be complex due to the presence of different types of hyperparameters (e.g. qualitative versus quantitative) and their inter-relationships (e.g. nested structure). 
The second difficulty is the large amount of time required for training to converge under a given configuration. 
%Nonconvex optimization, first-order and stochastic algorithms, and large data are among the contributing factors to the slow training phenomena. 
Due to these two difficulties, HPO has become rather an art than science, and practitioners have to rely on past experiences and intuitions to train and find a good model. Thus there is a huge demand for robust and effective methods that can automate HPO. 
%which is considered an indispensable component in a general effort to make ML automated (AutoML). 
Various HPO methods have been developed in the past ten years. Brute-force methods such as grid search and random search \cite{Bengio2011, Bergstra2012} were among the early works. A recently proposed method named  
%which can be paralleled but with drawback of computation inefficiency. Much recent progress has been made in HPO, like 
Hyperband \cite{Hyperband} uses the bandit idea to allocate more resources to promising configurations and terminate poor configurations. These methods directly compare different configurations without modeling $y$. Two model-based HPO methods are SMAC \cite{smac} and GP-BO \cite{Snoek2012}, both of which use Bayesian Optimization (BO) to construct prediction models and then sequentially find optimal configurations. SMAC uses random forests to predict the mean and variance of $y$ at $\bx$, and performs well with discrete hyperparameters. However, SMAC does not explicitly consider the correlation between adjacent configurations. On the other hand, GP-BO employs Guassian Process (GP) to model $y$; therefore, it considers correlations between different configurations. 
%model $y$ as a Gaussian distribution whose mean and variance are the empirical mean and variance over the predictions of the forest? trees \cite{smac, hutter13} and it performs well with  discrete or partial discrete hyperparameters. SMAC is a non-parametric method and does not consider the correlations between different hyperparameter configurations. While GP-BO uses a Gaussian process (GP) to model $y$ by allocating covariance structure on different configurations. 
Unlike SMAC and GP-BO that model $y$ directly, %Tree-structured Parzen Estimator (TPE) which is a non-standard Bayesian optimization algorithm
Tree-structured Parzen Estimator (TPE) uses kernel density estimation (KDE) to approximate $p(y<y^*)$, $p(\bx|y<y^*)$ and $p(\bx| y>y^*)$ for HPO, where $y^*$ is a fixed quantile of the observed performances \cite{Bengio2011, hutter13}. %TPE does not require a specific model for $y$ because of using KDE. 
Some recent methods combine the bandit idea and the BO strategy to optimize $y(\cdot)$ \cite{Swersky2013, Kandasamy2016gp, Klein2017, BOHB}. For example, Falkner et al. (2018) \cite{BOHB} combines Hyperband and TPE in the BO framework to create a partially model-based Hyperband method called BOHB, which demonstrates top performances with robustness and flexibility. 

%\cite{Swersky2013, Klein2017} studies the correlations between different tasks to hone the optimization process. \cite{Kandasamy2016gp} models the different fidelities of the objective function independently and integrate the multi-fidelity data in the sequential optimization stage of BO. 

% Motivations of the proposed method
Those existing methods reviewed above mainly differ in two aspects: (i) Whether the method models $y$ and to what degree; and (ii) whether and how it uses the bandit idea of different allocation of resources. 
% BOHB is the only one that combines both ideas, though its modeling of $y$ is using KDE. 
%In summary, two major ideas are underlying most existing methods, which are the bandit idea of different allocation of resources (e.g. training time) and the idea of modeling $y$. Compared to the existing methods reviewed above, the proposed method in this paper is new for HPO. Both Hyperband and BOHB implement LT and HT runs, and the configurations that perform badly in the LT runs are discarded even though they contain much information of the training system. 
The current paper represents another attempt to address the two major aspects facing HPO, especially the second slow training phenomena. 
%Based on our own experiences in training ML and deep learning models, 
We observed that the training procedure under a given configuration usually including the fast-improving period and the slow-improving period. During the fast-improving period, the objective function steadily improves over time, whereas during the slow-improving period, the objective function improves at a much slower pace.
%and thus the name for this period. This period is immediately followed by the slow-improving period, during which the objective function improves in a drastically slower pace. 
The fast-improving period usually achieves up to 90\% of the total improvement in the objective function in up to 20\% of the total training time, whereas the slow-improving period uses 80\% of the time to achieve the rest 10\% improvement. This observation suggests that the slow-training difficulty is mainly caused by the slow-improving period. In order to mitigate the slow training difficulty for HPO, we suggest to perform two types of training. The first type is called \emph{heavy training} (HT), which is to train the model under a given configuration to its completion, that is a certain convergence criterion has been met or the performance cannot be further improved. The second type is called \emph{light training} (LT), which is to end the model training right after the fast-improving period ends. Light training can be considered as an early stopping strategy. Various methods can be applied to detect the transition from the fast-improving period to the slow-improving period. 
%One example is to generate the screen plot and use the elbow-point of the plot as the early stopping time or just use small number of iterations instead of large number of iterations. (Can discuss use the number of iterations as an early stopping criterion, which can be learned from a few HT runs). 

LT runs are clearly much cheaper than HT runs in terms of computational cost and other resources. Although LT runs produce \emph{approximate measurements} of the performances of hyperparameter configurations, they are expected to be close to the \emph{accurate measurements}, and thus can be calibrated by a small number of judicially chosen HT runs. The strategy of using more accurate measurements to calibrate less accurate measurements for economical model building is related to the modeling of multi-fidelity computer experiments \cite{Kennedy2001, BHM, Xiong2013, He_Tuo_2014, Ezzat2018}. Those methods integrate low-fidelity and high-fidelity computer experiments via several Gaussian process models. In the multi-fidelity computer experiments, outcomes with different levels of fidelities are usually unordered. While stochastic orders sometimes exist between the approximate measurements and the accurate measurements in the HPO problem studied here. Motivated by the work of Kennedy and O’Hagan (2001) \cite{Kennedy2001} and Qian and Wu (2008) \cite{BHM}, we propose to integrate LT runs and HT runs via truncated Gaussian process models to capture the order information contained in the two levels of measurements of performances for the HPO purpose. Furthermore, we design a sequential strategy in BO framework for scoring candidate configurations for different levels of training. Compared with existing HPO methods mentioned above, our method combine the bandit idea and the fully modeling of the two levels of measurements. Therefore, our method not only efficiently finds optimal configurations, it also estimates the full performance profile of the configuration space.

% paper structure
We organize the rest of the article as follows. In Section \ref{sec:MethodsReview}, we briefly review the Bayesian optimization and Gaussian process. We propose a statistical model to systematically integrate the LT and HT data, and develop a new HPO algorithm in Section \ref{sec:ProposedMethods}. We apply the proposed method to some synthetic functions as well as some machine learning problems, and present the results in Section \ref{sec:experiments}. We conclude this article with a discussion in Section \ref{sec:conclusion}.

\section{Reviews of some relevant works} 
\label{sec:MethodsReview}

%Existing Bayesian Optimization (BO) methods only focus on modeling accurate measurements for HT runs and ignore approximate measurements for HT runs, but when LT runs are much more than HT runs, LT runs can provide much information. 

In this section, we briefly review the essentials of the Bayesian optimization. Further we provide the principal ideas of the Gaussian process model.

%and the design strategy[we should address the strategy previously] named nested Latin hypercube designs (NLHD) \cite{NLHD} to sample the configurations at two levels. used in single-fidelity computer experiments

\subsection{Bayesian optimization}

One major problem in computer experiments is to find the input that can minimize an objective function of interest, i.e., $\min_{\bx\in\cX} y(\bx)$. Bayesian optimization (BO) is a powerful tool that can resolve this kind of optimization problem with a key two-stage idea \cite{Jones1998, Shahriari2016}. In machine learning algorithms, the hyperparameter configuration $\bx$ and the accurate measurement of performance $y(\bx)$ obtained by the HT run at $\bx$ can be treated as the input and the objective function, respectively. Thus, the HPO problem can be resolved in the classic BO framework as discussed in Snoek et al. (2012) \cite{Snoek2012}. In the first stage of BO, a surrogate model is built to approximate $y(\bx)$. Based the prediction model obtained from the first stage, a criterion named \emph{acquisition function} is constructed to score and select new candidate configurations in the second stage.

%optimizing an Acquisition Function (AF) inherited from the first stage to perform the trade off between the exploration and exploitation \cite{Jones1998, Brochu2010A}. 

\subsection{Gaussian processes}\label{sec:GP}

%A common choice of the surrogate model is the Gaussian process model is commonly used for modeling have been extensively discussed	

Gaussian process model is a common choice as a surrogate for modeling $y$ with a stochastic process, which has been extensively studied in Santner et al. (2003) \cite{SWN03}.
%which is an infinite dimensional extension of multivariate Gaussian distribution \cite{Rasmussen2004}. 
For any two $d$-dimensional configurations $\bx=(x_1,\ldots,x_d)$ and $\bx'=(x'_1,\ldots,x'_d)\in\cX\subseteq \cR^d$, $y(\bx)$ is modeled as a Gaussian process as follows:
\begin{eqnarray}\label{eq:GP}
\bbE(y(\bx))=\mu,~~ \Cov(y(\bx), y(\bx')) = \sigma^2 R_{\bphi}(\bx-\bx'),
\end{eqnarray}
where $\mu$ and $\sigma^2$ are the mean and variance of the Gaussian process model, and $R_{\bphi}(\cdot)$ is the correlation function with length-scale  parameters $\bphi = (\phi_1,...,\phi_d)$. The Gaussian kernel is often used as the correlation function \cite{SWN03, Rasmussen2004} defined as
\begin{equation}\label{eq:CorrFunction}
R_{\bphi}(\bx-\bx') = \prod_{i=1}^d \exp\{-\phi_i(x_i-x_i')^2 \}.
\end{equation}
Thus, we denote $y(\cdot)\sim \GP(\mu, \sigma^2, \bphi)$, with unknown parameters $\btheta=(\mu,\sigma^2,\bphi)$. For clarity, the parameters in GP are not the hyperparameters we study in the machine learning algorithms. 

Suppose $D=\{\bx_1,\ldots, \bx_n\}$ is the set of configurations, with corresponding accurate measurements $\by=(y(\bx_1),\ldots,y(\bx_n))^T$. Let $\bR_{\bphi}$ denote the $n\times n$ correlation matrix of $\by$ whose $(i, j)$ element is $R_{\bphi}(\bx_i-\bx_j)$ as defined in \eqref{eq:CorrFunction}, and $\mathbf{1}_n$ is the $n$-dimensional column vector whose elements are all equal to 1. The expression of the log-likelihood of $\by$ is given by:
\begin{equation}\label{eq:log-likelihood}
l(\btheta) \propto -\frac{1}{2\sigma^2}(\by-\mu\cdot \mathbf{1}_n)^T\bR_{\bphi}^{-1}(\by-\mu\cdot \mathbf{1}_n)-\frac{n}{2}\ln(\sigma^2)-\frac{1}{2}\ln|\bR_{\bphi}|.
\end{equation}
By maximizing $l(\btheta)$, we have $\hat{\btheta}=(\hat{\mu},\hat{\sigma}^2,\hat{\bphi})$ (and thus, $\bR_{\hat{\bphi}}$) as the maximal likelihood estimation of $\btheta=(\mu,\sigma^2,\bphi)$, and let $\hat{\bR}=\bR_{\hat{\bphi}}$. Using the Bayes rule, the (posterior) predictive distribution of $y(\bx)\mid\by,\hat{\btheta}$ is Gaussian \cite{Jones1998, Olivier2012}: 
\begin{equation}\label{eq:PosteriorGP}
y(\bx)\mid \by,\hat{\btheta} \sim N\big(\hat{y}(\bx),s^2(\bx) \big),
\end{equation}
where
\begin{eqnarray}\label{eq:GP_PosteriorMeanVar}
\hat{y}(\bx)&=&\hat{\mu} + \hat{\br}^T\hat{\bR}^{-1}(\by-\hat{\mu}\cdot\mathbf{1}_n), \nonumber \\
s^2(\bx)&=& \hat{\sigma}^2(1-\hat{\br}^T\hat{\bR}^{-1}\hat{\br}),
\end{eqnarray}
with $\hat{\br}$ representing the correlation vector between $y(\bx)$ and $\by$ with the estimated length-scale parameters $\hat{\bphi}$.

\begin{comment}
Given $\bphi$, the MLE of $\mu$ and $\sigma^2$ are 
\begin{eqnarray}\label{eq:MLE-mu-sigma}
\hat{\mu}=\frac{\mathbf{1}_n^T \bR^{-1}\by}{\mathbf{1}_n^T \bR^{-1}\mathbf{1}_n},~~ \hat{\sigma}^2=\frac{1}{n}(\by-\hat{\mu}\cdot \mathbf{1}_n)^T\bR^{-1}(\by-\hat{\mu}\cdot \mathbf{1}_n).
\end{eqnarray}
Plugging $\hat{\mu}$ and $\hat{\sigma}^2$ in \eqref{eq:log-likelihood}, the MLE of $\bphi$ is obtained as $\hat{\bphi}=\argmax_{\bphi}[-n\ln(\hat{\sigma}^2)-\ln|R|]$. The estimation of $\bR$ is denoted as $\hat{\bR}$ whose $(i, j)$ element is $R(\bx_i-\bx_j\mid \hat{\bphi})$. Using the Bayes rule, the distribution of $y(\bx)$ conditioned on $\by$ is a Gaussian distribution \cite{Jones1998, Olivier2012}, 
\begin{equation}\label{eq:PosteriorGP}
y(\bx)\mid \by \sim N\big(\hat{y}(\bx),s^2(\bx) \big),
\end{equation}
where
\begin{eqnarray}
\hat{y}(\bx)&=&\hat{\mu} + \hat{\br}^T\hat{\bR}^{-1}(\by-\hat{\mu}\cdot\mathbf{1}_n),\\
s^2(\bx)&=& \hat{\sigma}^2(1-\hat{\br}^T\hat{\bR}^{-1}\hat{\br}),
\end{eqnarray}
with $\hat{\br}$ representing the correlation vector between $y(\bx)$ and $\by$.
\end{comment}

\subsection{Acquisition function}

Based on the results provided in \eqref{eq:GP_PosteriorMeanVar}, BO seeks the next candidate configuration at the second stage by optimizing an acquisition function. A popular choice of the acquisition function is the \emph{Upper Confidence Bound} (UCB) which is given as follows	
%Expected Improvement (EI) \cite{Jones1998}. Considering the cost of fitting a model in the first stage, a sequential batched version can be applied in the second stage \cite{qEI, Wu2016}.
\begin{equation}\label{eq:UCB}
\text{UCB}(\bx) = -\hat{y}(\bx) + \beta_n s(\bx).
\end{equation}
Srinivas et al. (2010) \cite{Srinivas2010} suggested that setting $\beta_n$ as $0.2d\log(2n)$ can guarantee a good convergence rate. The UCB criterion quantifies the improvement made by a new configuration $\bx$, and it balances the exploration and exploitation under the surrogate model obtained so far \cite{Jones1998}. The next candidate configuration $\bx_{n+1}$ can be chosen as  
\begin{equation}\label{eq:MaxUCB}
\bx_{n+1} = \argmax_{\bx\in\cX} \text{UCB}(\bx).
\end{equation}		
The pseudocode for the GP-BO algorithm with the UCB acquisition function is provided in Algorithm \ref{alg:BO}.

\begin{algorithm}	
	\caption{Pseudocode for GP-BO algorithm}
	\label{alg:BO}
	\textbf{Input}: Maximal number of HT runs $N_{\text{max}}$\\
	\textbf{Initialization}: Generate $n$ configurations $D=\{\bx_1,\ldots, \bx_n\}$ and get the corresponding accurate measurements of performances $\by=\{y(\bx_1),\ldots,y(\bx_n)\}$
	\begin{algorithmic}[1]
		\For {$i=1,..., N_{\text{max}}$}
		\State Fit a GP model to obtain $\hat{y}(\bx)$ and $s(\bx)$
		\State Choose $\bx_{n+1} = \argmax_{\bx\in\cX}\text{UCB} (\bx)$
		\State Evaluation $y(\bx_{n+1})$, and set $D= D \cup \{\bx_{n+1}\}, \by = \by \cup \{y(\bx_{n+1})\}$		
		\EndFor 
		\State \textbf{return} $\bx^* = \argmin_{\bx \in D} \by$
	\end{algorithmic}
\end{algorithm}

\section{Proposed methods}
\label{sec:ProposedMethods}

In this section, we propose the \emph{Truncated Additive Model} (TAM) to fit the 
accurate measurements of performance and develop a novel algorithm to find the optimal hyperparameter configuration in the BO framework.

\subsection{Truncated Additive Model}
\label{sec:TAM}

Let $y_l(.)$ denote the approximate measurements of performance generated by LT runs, and $y_h(.)$ denote the accurate measurements of performance generated by HT runs. In most machine learning algorithms, there exists stochastic orders between $y_l(\cdot)$ and $y_h(\cdot)$. For example, given one configuration $\bx$ of a specific machine learning algorithm, the validation error $y_h(\bx)$ reported by the HT run is less than the approximate validation error $y_l(\bx)$ reported by the LT run. More generally, we propose the \emph{Truncated Additive Model} (TAM) to integrate $y_l$ and $y_h$ for any $\bx\in\cX$:
\begin{eqnarray}\label{eq:TAM}
y_h(\bx) &=& \rho y_l(\bx) + \delta(\bx), \nonumber \\
y_l(\bx) &\sim& \GP(\mu_l, \sigma_l^2, \bphi_l), \nonumber \\
\delta(\bx)&\sim&\GP(\mu_{\delta}, \sigma_{\delta}^2, \bphi_{\delta})\cdot \bbI(\delta_1\leq \delta(\bx) \leq\delta_2), ~\delta_1\leq \delta_2\in\bbR,
\end{eqnarray}
where $\bbI(\cdot)$ is the indicator function. We assume $y_l(\cdot)$ to be a Gaussian process, $\delta(\cdot)$ to be a truncated Gaussian process with a truncated interval $[\delta_1, \delta_2]$ which represents the location adjustment, $\rho$ to be an unknown parameter measures the scale change from LT run to HT run, and $y_l(\cdot)$ and $\delta(\cdot)$ to be independent. The proposed model in \eqref{eq:TAM} can be viewed as an extension of \cite{Kennedy2001} and \cite{BHM}, where $\delta(\cdot)$ is a Gaussian process. Considering the validation error in a machine learning algorithm, we can set $\rho\leq 1$ and $\delta_2\leq 0$ to satisfy $y_h(\bx)\leq y_l(\bx)$.

\subsection{Statistical inference}\label{sec:StatInferOfTAM}

Let $D_l=\{\bx_1^l,\ldots, \bx_{n}^l\}$ denote the configuration set for $n$ LT runs, $\by_l=(y_l(\bx_1^l),.., y_l(\bx_n^l))^T$ denote the corresponding approximate measurements, $D_h=\{\bx_1^h,\ldots, \bx_{n_1}^h\}$ denote the configuration set for $n_1$ HT runs, and $\by_h=(y_h(\bx_1^h),...,$ $y_h(\bx_{n_1}^h))^T$ denote the corresponding accurate measurements. In machine learning algorithms, we always obtain the accurate measurement after the approximate measurement, and thus $D_h\subset D_l$. Let $\bR_l=\bR_{\bphi_l}$ and $\bR_{\delta}=\bR_{\bphi_{\delta}}$ be the correlation matrix of $\by_l$ and $\bdelta=[\delta(\bx_1^h),\ldots,\delta(\bx_{n_1}^h)]^T$, respectively. Given the unknown parameters $\btheta = (\rho, \mu_l, \mu_{\delta}, \sigma_l^2, \sigma_{\delta}^2, \bphi_l, \bphi_{\delta})$, the distribution of $\by_h|~\by_l,\btheta$ is a multivariate truncated normal (TN) distribution:
\begin{eqnarray}
\by_h|~\by_l,\btheta \sim \text{TN}_{n_1}(\rho\by_{l_1}+\mu_{\delta}\bm{1}_{n_1}, \sigma_{\delta}^2 \bR_{\delta}; ~\rho \by_{l_1}+\delta_1 \bm{1}_{n_1} , \rho \by_{l_1}+\delta_2 \bm{1}_{n_1}), 
\end{eqnarray}
where $\by_{l_1}=(y_l(\bx_1^h),\ldots,y_l(\bx_{n_1}^h))^T$. And $\bX\sim \text{TN}_n(\ba,\bB; \bc,\bd)$ stands for the distribution of the $n$-dimensional normal vector $\bX$ conditional on $\bc\leq\bX\leq \bd$, where $\ba$ and $\bB$ are the mean vector and covariance matrix of $\bX$, respectively. Then the density function of $\by_h|~\by_l,\btheta$ is given by
\begin{equation*}
f_{\by_h|\by_l, \btheta}(\by_h) = \frac{\exp\{-\frac{1}{2}(\by_h-\rho\by_{l_1}-\mu_{\delta}\bm{1}_{n_1})^T (\sigma_{\delta}^2\bR_{\delta})^{-1}(\by_h - \rho\by_{l_1}-\mu_{\delta}\bm{1}_{n_1})\}}{Z_h(\rho, \mu_{\delta}, \sigma_{\delta}^2, \bphi_{\delta})},
\end{equation*}
where $$Z_h(\rho, \mu_{\delta}, \sigma_{\delta}^2, \bphi_{\delta})=\int_{\by_h\in R_{n_1}} \exp\{-\frac{1}{2}(\by_h-\rho\by_{l_1}-\mu_{\delta}\bm{1}_{n_1})^T (\sigma_{\delta}^2\bR_{\delta})^{-1}(\by_h - \rho\by_{l_1}-\mu_{\delta}\bm{1}_{n_1})\} d\by_h$$
and $R_{n_1}=[\rho\by_{l_1}+\delta_1\bm{1}_{n_1}, \rho\by_{l_1}+\delta_2\bm{1}_{n_1}]$. The log-likelihood of $(\by_h,\by_l)$ can be shown to be
\begin{eqnarray}\label{eq:TAM-MLE}
l(\btheta) &=& \log [p(\by_h |~\by_l, \btheta)p(\by_l |~\btheta)] \nonumber \\
&\propto&-\frac{1}{2\sigma_{\delta}^2}(\by_h-\rho\by_{l_1}-\mu_{\delta}\bm{1}_{n_1})^T \bR_{\delta}^{-1}(\by_h-\rho\by_{l_1}-\mu_{\delta}\bm{1}_{n_1})-\frac{n}{2}\ln \sigma_l^2-\frac{1}{2}|\bR_l| \nonumber \\
&& -\frac{1}{2\sigma_l^2}(\by_l-\mu_l\bm{1}_{n})^T \bR_l^{-1}(\by_l-\mu_l\bm{1}_{n})-\ln \{Z_h(\mu_{\delta}, \sigma_{\delta}^2, \bphi_{\delta})\}, 
\end{eqnarray}
where $\bR_l$ is the correlation matrix of $\by_l$. It is easy to check that $l(\btheta)$ take its maximum at 
$$\hat{\mu}_l=\frac{\bm{1}_{n}^T \bR_l^{-1}\by_l}{\bm{1}_{n}^T \bR_l^{-1}\bm{1}_{n}}, ~~\hat{\sigma}^2_l=\frac{1}{n}(\by_l-\hat{\mu}_l\bm{1}_{n})^T \bR_l^{-1}(\by_l-\hat{\mu}_l\bm{1}_{n}).$$
Plugging $\hat{\mu}_l$ and $\hat{\sigma}^2_l$ into \eqref{eq:TAM-MLE},
the MLE $(\hat{\rho},\hat{\mu}_{\delta},\hat{\sigma}_{\delta}^2, \hat{\bphi}_l, \hat{\bphi}_{\delta})$ of $(\rho,\mu_{\delta},\sigma_{\delta}^2, \bphi_l, \bphi_{\delta})$ can be obtained by numerical optimization algorithms such as L-BFGS \cite{Nocedal1980}. We denote the MLE of $\btheta$ as $\hat{\btheta}$.

For any $\bx\in D_l\setminus D_h$, $y_h(\bx),\by_h|~\by_l,\hat{\btheta}\sim \text{TN}_{n_1+1}(\hat{\rho}\by_{l_1}^*+ \hat{\mu}_{\delta}\bm{1}_{n_1}, M_{\delta}; \hat{\rho}y_l(\bx)+\delta_1, \hat{\rho}y_l(\bx)+\delta_2)$, and its
density function takes the form
\begin{eqnarray*}
	p(y_h(\bx),\by_h|~ \by_l, \hat{\btheta})=\frac{\exp\{-\frac{1}{2}(\by_h^*-\hat{\rho}\by_{l_1}^*- \hat{\mu}_{\delta}\bm{1}_{n_1})^T M_{\delta}^{-1}(\by_h^* - \hat{\rho}\by_{l_1}^*-\hat{\mu}_{\delta}\bm{1}_{n_1})\}}{\int_{y_h(\bx)\in R_1} \exp\{-\frac{1}{2}(\by_h^*-\hat{\rho}\by_{l_1}^*-\hat{\mu}_{\delta}\bm{1}_{n_1})^T M_{\delta}^{-1}(\by_h^* - \hat{\rho}\by_{l_1}^*-\hat{\mu}_{\delta}\bm{1}_{n_1})\} dy_h(\bx)}
\end{eqnarray*}
where $\by_{l_1}^*=(y_l(\bx), \by_l)^T$, $\by_h^*= (y_h(\bx), \by_h)^T, M_{\delta}=\sigma_{\delta}^2 \bR_{\delta}^*$, $R_1=[\hat{\rho}y_l(\bx)+\delta_1,\hat{\rho}y_l(\bx)+\delta_2]$, and $\bR_{\delta}^*$ is the $(n_1+1)\times (n_1+1)$ correlation matrix of $\by_h^*$ with estimated length-scale parameters $\hat{\bphi}_{\delta}$. Using the results from Horrace (2005) \cite{Horrace2005} and the Bayes rule 
$$p(y_h(\bx)|~\by_l, \by_h, \hat{\btheta}) = \frac{p(y_h(\bx),\by_h|~ \by_l, \hat{\btheta})}{p(\by_h|~ \by_l, \hat{\btheta})},$$ 
we have the following proposition that $y_h(\bx)|~\by_l, \by_h, \hat{\btheta}$ is truncated normal.	
\begin{proposition}\label{prop:TGP}
	\begin{eqnarray}
	y_h(\bx)|~\by_l, \by_h, \hat{\btheta}\sim \textup{TN}(\hat{\mu}_h(\bx), \hat{\sigma}_h^2(\bx); ~ \hat{\rho}y_l(\bx)+\delta_1, \hat{\rho}y_l(\bx)+\delta_2),
	\end{eqnarray}
	where
	\begin{eqnarray*}
		\hat{\mu}_h(\bx)&=& \hat{\rho} y_l(\bx)+\hat{\mu}_{\delta}+\hat{\br}_{\delta}^T \hat{\bR}_{\delta}^{-1}(\by_h-(\hat{\rho}\by_l+\hat{\mu}_{\delta}\bm{1}_{n_1})),\\ \hat{\sigma}_h^2(\bx)&=& \hat{\sigma}_{\delta}^2(1-\hat{\br}_{\delta}^T \hat{\bR}_{\delta}^{-1}\hat{\br}_{\delta}),
	\end{eqnarray*}
	$\hat{\bR}_{\delta}$ is the correlation matrix of $\bdelta$ whose $(i,j)$ element is $R_{\hat{\bphi}_{\delta}}(\bx_i^h-\bx_j^h)$, and $\hat{\br}_{\delta}$ is the correlation vector between $\delta(\bx)$ and $\bdelta$ with estimated length-scale parameters $\hat{\bphi}_{\delta}$.	
\end{proposition}	

Johnson et al. (1994) \cite{Johnson1994} has provided the mean and variance of the truncated normal random variable. Therefore, we can directly derive the mean and variance of of $y_h(\bx)|~\by_l, \by_h, \hat{\btheta}$ and show them as follows:
\begin{eqnarray}
\hat{y}_h(\bx) &=& \hat{\mu}_h(\bx) + \frac{\phi(\alpha)-\phi(\beta)}{Z}\hat{\sigma}_h(\bx),\\ 
s_h^2(\bx) &=& \hat{\sigma}_h^2(\bx) \Big[1 + \frac{\alpha\phi(\alpha)-\beta\phi(\beta)}{Z} - \Big(\frac{\phi(\alpha)-\phi(\beta)}{Z}\Big)^2 \Big],
\end{eqnarray}
where 
$$\alpha=\frac{\hat{\rho}y_l(\bx)+\delta_1-\hat{\mu}_h(\bx)}{\hat{\sigma}_h(\bx)} ~\beta=\frac{\hat{\rho}y_l(\bx)+\delta_2-\hat{\mu}_h(\bx)}{\hat{\sigma}_h(\bx)},~ Z = \Phi(\beta) - \Phi(\alpha),$$ 
and $\phi(\cdot)$ and $\Phi(\cdot)$ are the density function and cumulative distribution function of the standard normal distribution.

For any $\bx\in\cX$, we consider two prediction scenarios:
\begin{itemize}
	\item[1.] When $\bx_0 \in D_l \setminus D_h$, the prediction of $y_h(\bx_0)$ can be chosen as $\hat{y}_h(\bx_0)$.
	\item[2.] When $\bx_0 \notin D_l$, $y_l(\bx_0)$ is missing. From \eqref{eq:PosteriorGP}, $y_l(\bx_0) | \by_l, \hat{\btheta}$ has a normal distribution with the mean $\hat{y_l}(\bx_0)$ and variance $s_l(\bx_0)$. We can impute $y_l(\bx_0)$ by $\hat{y_l}(\bx_0)$ so that $y_h(\bx_0)$ can be predicted as in the first scenario.
\end{itemize} 

%Consider the LT loss $y_l(\bx)$ and HT loss $y_h(\bx)$ in ML algorithms, we can simplify the model by setting $\rho=1$ and $\delta_2\leq 0$ to satisfy $y_h(\bx) \leq y_l(\bx)$.

\subsection{Bayesian truncated additive optimization (BTAO)} \label{sec:BTAO}

In this subsection, we will introduce a new HPO method called \emph{Bayesian Truncated Additive Optimization} (BTAO) algorithm to find the optimal hyperparameter configuration. BATO is based on the BO framework with two key stages as introduced in Section \ref{sec:MethodsReview}. Previously we have used TAM to establish the surrogate model to integrate LT and HT runs for the first stage. We then introduce how to design a strategy to choose the candidate configurations for the second stage.

We notice that evaluating LT runs is not so costly. On the other hand, the configurations that perform well in LT runs have potentials to improve the performances in HT runs. Therefore, we can design a strategy to select a batch of candidate configurations for LT runs and judiciously select one configuration for HT run among those candidates. We define the acquisition function for LT run as
\begin{equation}
\text{UCB}_l(\bx) = -\hat{y}_l(\bx) + \beta_n s_l(\bx),
\end{equation}
and the acquisition function for HT run as
\begin{equation}
\text{UCB}_h(\bx) = -\hat{y}_h(\bx) + \beta_{n_1} s_h(\bx).   
\end{equation}
We will explain and summarize the procedure in Algorithm \ref{alg:bato}.

As for the initialization, we recommend to use the nested Latin hypercube designs (NLHD) \cite{NLHD, Xiong2013} to generate a pair of collections of configurations $(D_l, D_h)$. This procedure is to guarantee the two levels of configurations have space-filling properties. Then updating the Gaussian process model based on LT data $(D_l,\by_l)$ and sequentially add $s$ ($s>1$) configurations for LT runs (line 2-6). For each $j$ in the inner loop, $\bx_{l, i_j}$ is obtained by maximizing $\text{UCB}_l(\bx)$ over the configuration space $\cX$ (line 4). This manipulation can find the configurations that have potentials to improve the performance in LT runs.  
%On the other hand, unlike fitting a single GP model on LT data, it is more complex and time-consuming to fit a TAM on LT and HT data. 
After that, we choose the next candidate configuration $\bx_{h, i}$ for HT run from $D_l\setminus D_h$ (line 8). As the optimization progresses, both the LT runs and HT runs can balance between the exploitation and exploration. 
%The theorem in \cite{NLHD, Stein198787} guarantees that BHM can fit the full performance profile with less variation.

\begin{algorithm}
	\caption{Pseudocode for BTAO}
	\label{alg:bato}
	\textbf{Input}: Maximal number of HT runs $N_{\text{max}}$\\
	\textbf{Initialization:} Generate LT and HT configurations $(D_l, D_h)$ using NLHD satisfying $|D_l|=s|D_h|$, and evaluate the corresponding validation errors $(\by_l,\by_h)$
	\begin{algorithmic}[1]
		\For {$i=1,\ldots,N_{\text{max}}$}
		\For {$j=1,\ldots,s$}
		\State  Fit a GP model based on LT data $(D_l,\by_l)$	
		%$(\hat{y}_{l, i_j}, s_{l, i_j},  \hat{y}_{h, j}, s_{h, i_j}) = \text{TAM}(D_l, \by_l,D_h, \by_h)$ 
		\State $\bx_{l, i_j} = \argmax_{\bx\in\cX}\text{UCB}_l (\bx)$
		\State Get the light training evaluation $y_{l, i_j}$ on  $\bx_{l, i_j}$, and set $D_l = D_l \cup \{\bx_{l, i_j}\}, \by_l = \by_l \cup \{y_{l, i_j}\}$
		\EndFor 	
		\State  Fit a TAM based on LT and HT data $(D_l, \by_l,D_h, \by_h)$		
		\State  Choose one configuration with the largest $\text{UCB}_h$ value from $D_l \setminus D_h$ as $\bx_{h, i}$, and get the heavy training evaluation $y_{h, i}$ on $\bx_{h, i}$. Set $D_h=D_h \cup \{\bx_{h, i}\}$ and $\by_h=\by_h \cup \{y_{h, i}\}$
		\EndFor 
		\State \textbf{return} $\bx^* = \argmin_{\bx \in D_h} \, \by_h$
	\end{algorithmic}
\end{algorithm}

\begin{figure}[H]
	\centering
	\includegraphics[width=4.5in]{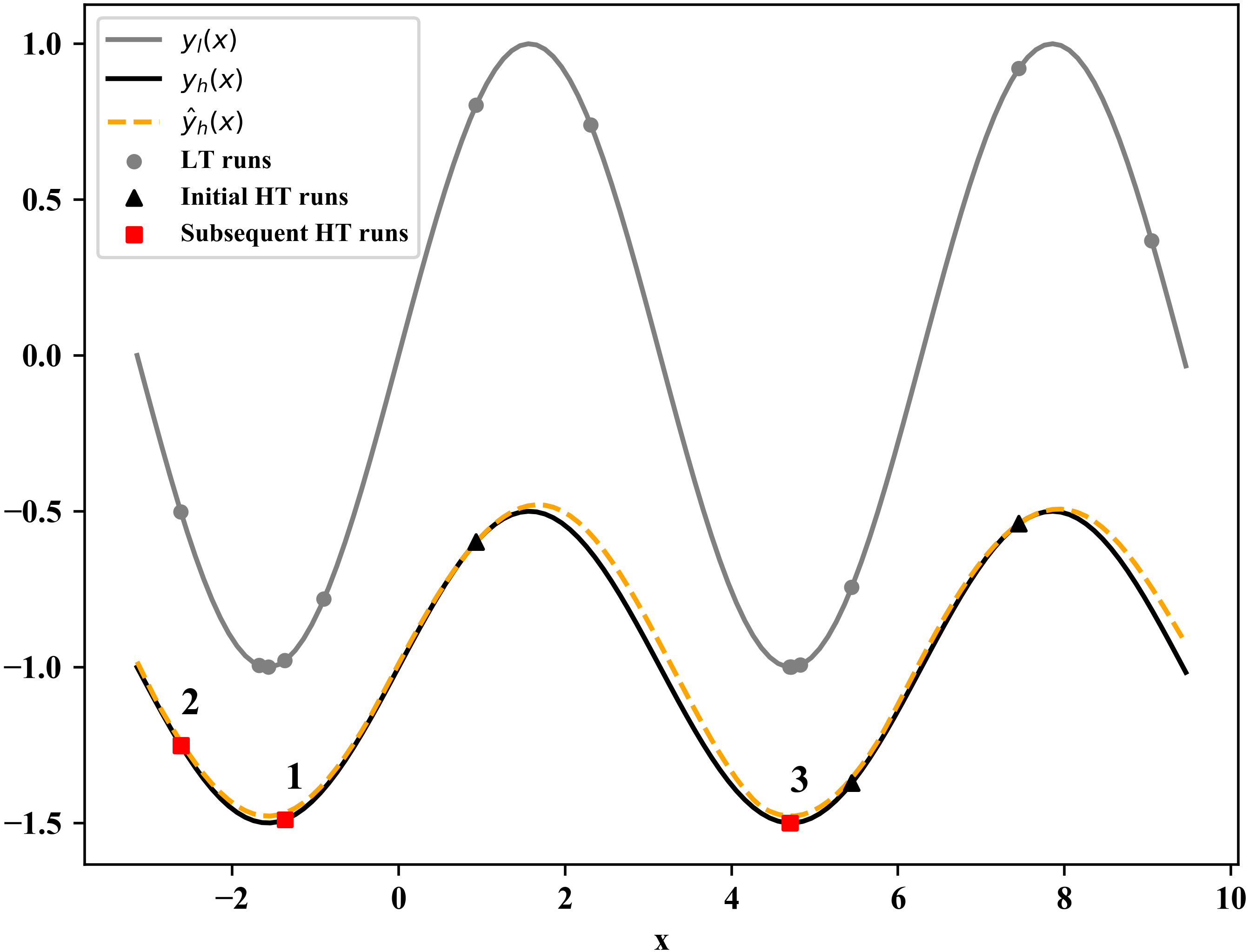}
	\caption{Finding the minima using BATO.}\label{fig:Sin}
\end{figure}

We present a toy example in Figure \ref{fig:Sin} for the illustration of the proposed BTAO approach, where $y_l(x)=\sin x$, $y_h(x) = 0.5y_l(x)-1$, $x\in [-\pi, 3\pi], \delta_1=-1.5,\delta_2=0.5$. In the plot, the solid lines in gray and black represent the true response curve of $y_h$ and $y_l$, respectively. We first use NLHD to generate three initial HT runs denoted by black-upper triangles. In the later iterations of BTAO, three additional HT runs are generated which are denoted by red squares, and the numbers above the squares indicate in which iteration they are generated. The initial and subsequent LT runs are all denoted by gray dots. In the third iteration, the orange dashed line is the predicted response curve of $y_h$ using TAM with 6 HT runs and 12 LT runs, which is very close to the true response curve of $y_h$. In this example, BATO can effectively find the minima in limited iterations.

\section{Experiments}\label{sec:experiments}

We first introduce some parameters that determine the approximate measurements of LT runs and accurate measurements of HT runs. Let $E_l$ and $E_h$ denote the maximal number of iterations (i.e., epochs in deep neural networks) for each LT run and HT run, $s_l$ and $s_h$ denote the length of training iteration strips, $e_l$ and $e_h$ denote the thresholds of measurement improvement in LT runs and HT runs, respectively. In LT runs, the approximate measurements are defined by a triple of parameters $(E_l, s_l, e_l)$. To be concrete, if one configuration reaches a maximal number of training iterations $E_l$ or has an improvement of the measurement which is less than $e_l$ in $s_l$ successive iterations, we early stop the training process and call it LT run to obtain the approximate measurement. By replacing $(E_l, s_l, e_l)$ with $(E_h, s_h, e_h)$, the accurate measurement for HT run can be defined in the same way. In this paper, we consider the validation error as the measurement of performance in ML or DL algorithms.

We evaluate the performances of BTAO in several tasks, including finding the maximum of two synthetic examples taken from Xiong et al. (2013) \cite{Xiong2013}, and the other three experiments: optimizing the hyperparameters of support vector machines, fully-connected neural networks, and convolutional neural networks. Five other methods including Random Search (RS), TPE, GP-BO, SMAC and BOHB are used for comparison.

\subsection{Synthetic functions}

The task in this subsection is to find the maximum of the synthetic function $y_h(\bx)$. Two synthetic examples, the Currin exponential example and Park example from Xiong et al. (2013) \cite{Xiong2013} are considered. 
%and each of them is associated with one LT function denoted by $y_l(\bx)$ and one HT function denoted by $y_h(\bx)$. 
Since BOHB requires more than two different levels of functions and the code provided by \cite{BOHB} does not work for the synthetic examples, BOHB is not applied to the two examples. Let $y_h^*$ denote the maximum of $y_h(\bx)$ over $\cX$. We define the \emph{simple regret}: $S_{n_1} = y_h^*-\max_{\bx\in D_h} y_h(\bx)$. The quantity $S_{n_1}$ can evaluate the performances of different methods over the number of HT function evaluations. Note that $\max_{\bx\in\cX} y_h(\bx)$ is equivalent to $\min_{\bx\in\cX} -y_h(\bx)$, so we model $-y_h(\bx)$ instead of $y_h(\bx)$ in the BTAO algorithm.

\subsubsection{Currin exponential example}

The domain of the Currin exponential example is the two dimensional unit cube $\cX=[0,1]^2$. The HT function is defined as
\begin{eqnarray*}
	y_h(\bx)&=&\Big[1-\exp\Big(-\frac{1}{2x_2}\Big)\Big] \frac{2300x_1^3+1900x_1^2+2092x_1+60}{100x_1^3+500x_1^2+4x_1+20},
\end{eqnarray*}
and the LT function is defined as
\begin{eqnarray*}
	y_l(\bx)&=&\frac{1}{4}y_h(x_1+0.05, x_2+0.05)+\frac{1}{4}y_h(x_1+0.05, \max(0,x_2-0.05))+\\
	&&\frac{1}{4}y_h(x_1-0.05, x_2+0.05)+\frac{1}{4}y_h(x_1-0.05, \max(0,x_2-0.05)).
\end{eqnarray*}

\subsubsection{Park example}
The domain of the Park example is $\cX=[0,1]^4$, with the HT function defined as
\begin{eqnarray*}
	y_h(\bx)&=&\frac{2}{3}\exp(x_1+x_2)-x_4\sin(x_3)+x_3,
\end{eqnarray*}
and the LT function defined as
\begin{eqnarray*}
	y_l(\bx)&=& 1.2y_h(\bx)-1.
\end{eqnarray*}

\begin{figure}[H]
	\subfigure[Currin exponential example $(d=2)$]{
		\label{fig:Currin}
		\includegraphics[width=2.3in]{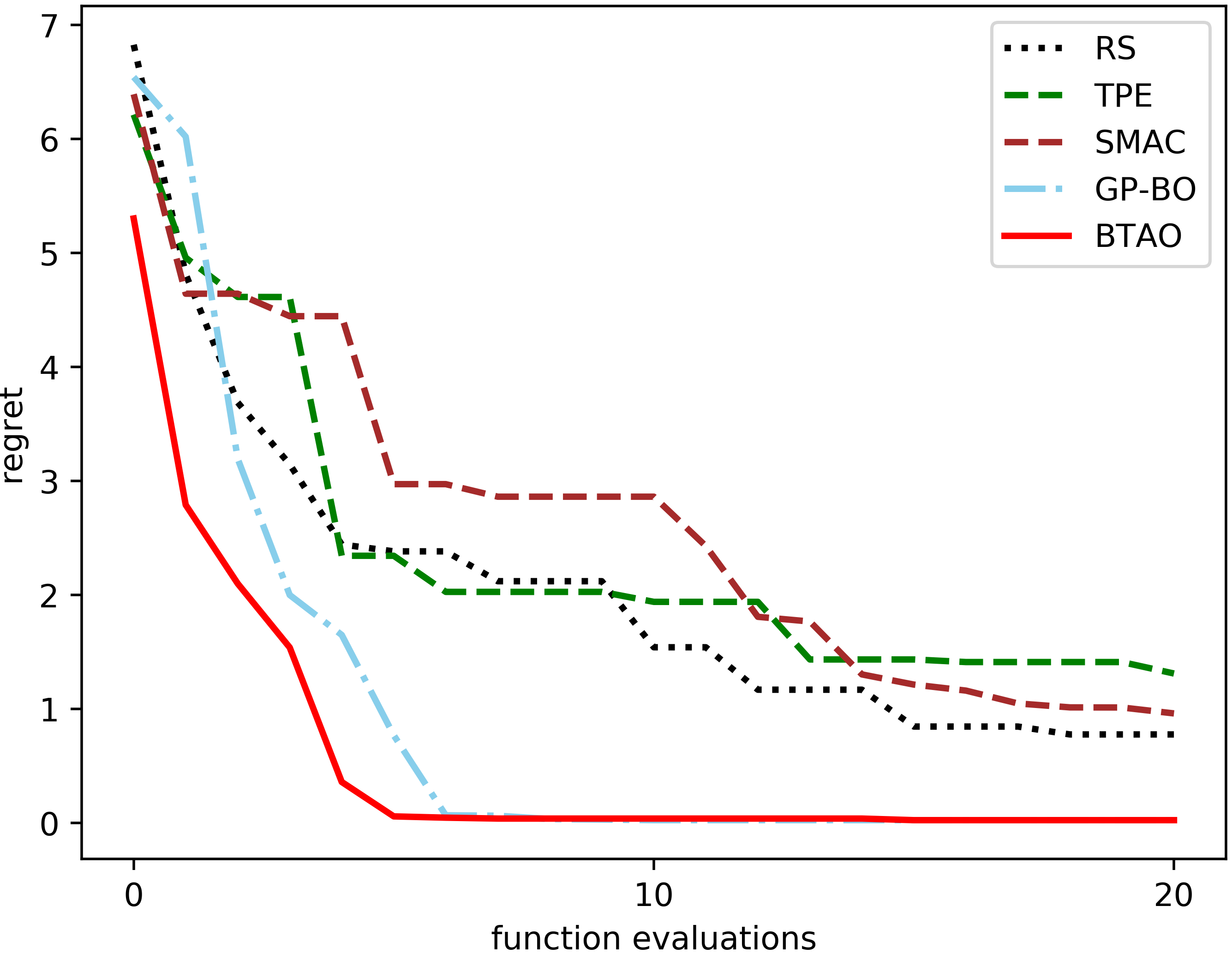}}
	\hspace{.05in}
	\subfigure[Park example $(d=4)$]{
		\label{fig:Park}
		\includegraphics[width=2.4in]{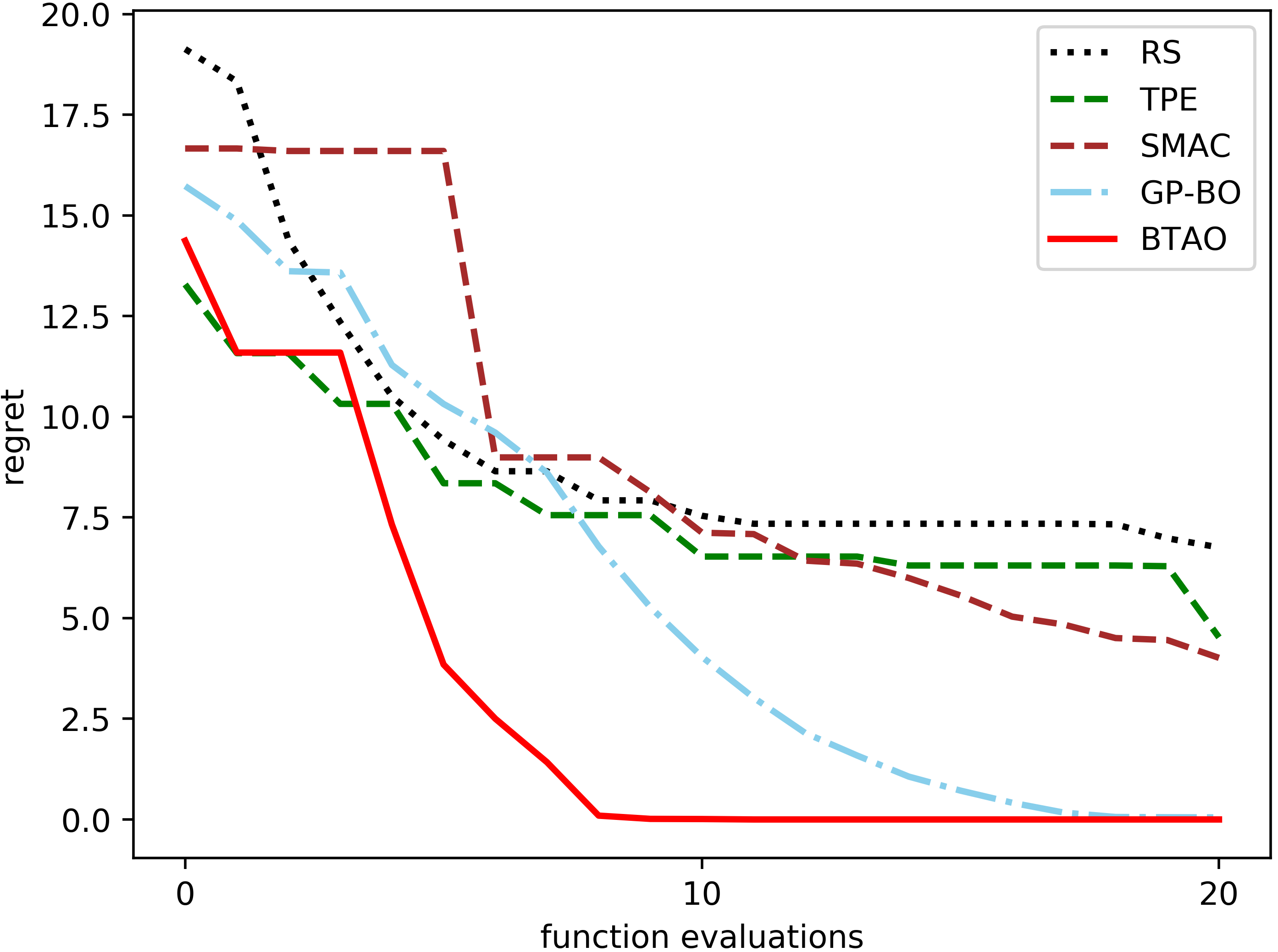}}
	\caption{The simple regret against the number of HT function evaluations. All curves are produced by averaging over 10 independent experiments.}
	\label{fig:SynExample}
\end{figure}

In Figure \ref{fig:SynExample}, BTAO outperforms other four single-fidelity methods on the two synthetic examples. Both GB-BO and BTAO can achieve the same best performance with zero simple regret in a limited number of HT function evaluations, but GB-BO takes more HT function evaluations especially when the dimension increases.

%TPE can achieve the same best performance with a zero simple regret. With the increase of the iterations, BTAO and SMAC had comparable performances and consistently performed better than Random Search, TPE and GP-BO. Although GP-BO was faster than RS and TPE, it was slower than BTAO due to the lack of extra LT data.

\begin{comment}
\subsection{Synthetic function: Park2 function}
The domain is $\cX=[0,1]^4$, the HT function is
\begin{eqnarray*}
y_h(\bx)&=&\frac{x_1}{2} \Big(\sqrt{1+(x_2+x_3^2)\frac{x_4}{x_1^2}}-1\Big)+(x_1+3x_4)\exp(1+\sin(x_3)),
\end{eqnarray*}
and the LT function is
\begin{eqnarray*}
y_l(\bx)&=&\Big(1+\frac{\sin(x_1)}{10}\Big)y_h(\bx)-2x_1^2+x_2^2+x_3^3+0.5.
\end{eqnarray*}
\end{comment}

\subsection{Support Vector Machines on MNIST}
\begin{figure}[H]
	\centering
	\includegraphics[width=3.9in]{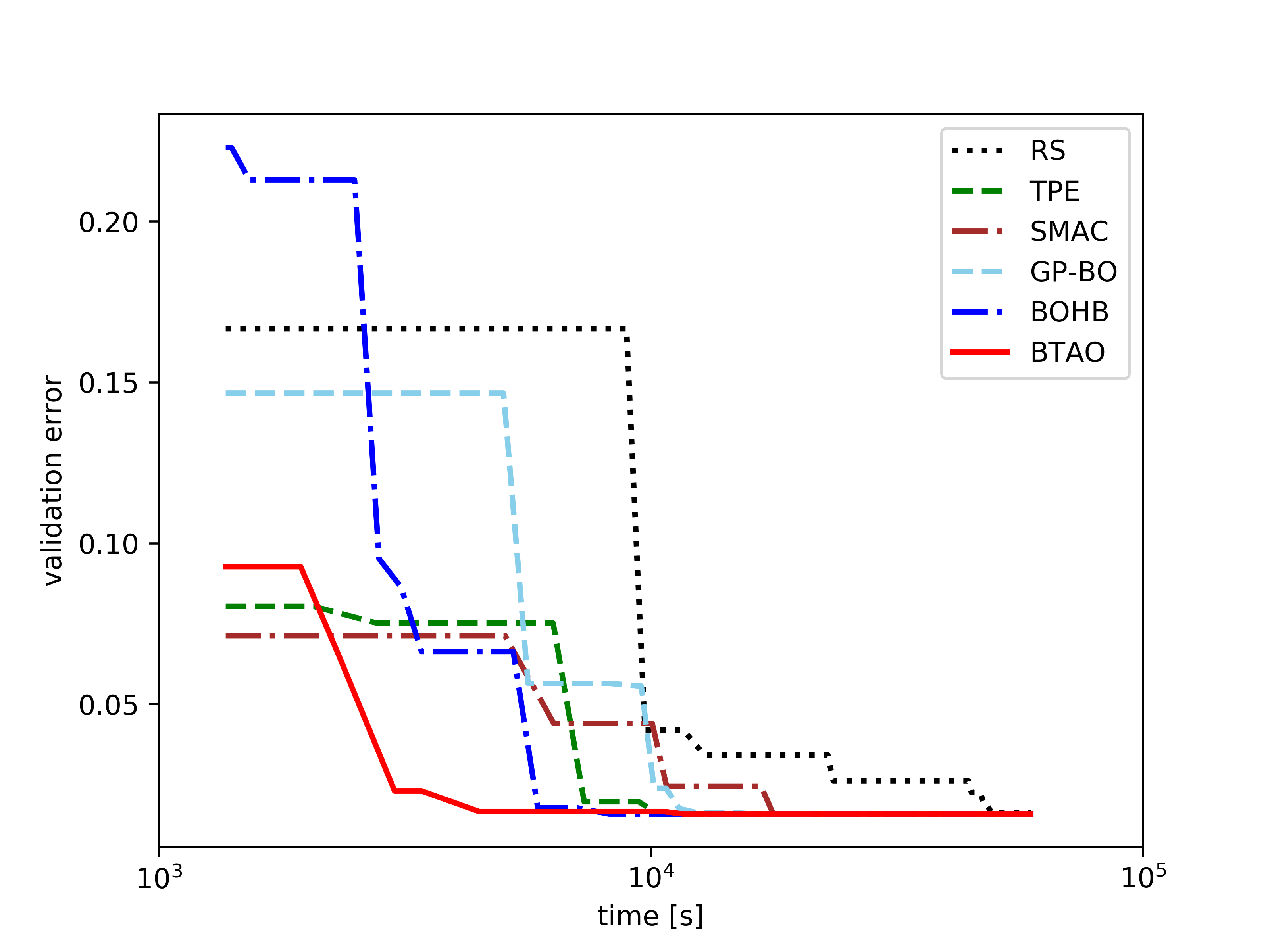}
	\caption{Optimizing two hyperparameters of SVM on MNIST.}
	\label{fig:SVM}
\end{figure}

%We first applied our method BTAO and other methods to optimize the hyperparameters of SVM on MNIST \cite{BOHB, Klein2017}. We used the linear kernel SVM, and only one hyperparameter that is the penalty factor $C$ needs to be tuned, and $C \in [2^{-10}, 2^{10}]$. For MNIST, we used 50,000 images as the training set and 10,000 images as the validation set, $E_l=300$ iterations for LT runs and $E_h=1000$ iterations for HT runs. As Figure \ref{SVM_1} (b) shows, Random Search, SMAC, GP-BO, BOHB and BATO have comparative final performances. TPE is a little bit worse than other 5 methods. As for the two fully model-based approaches GP-BO and BTAO, they both can generate the performance profile of all $C$ in its range. In this simple one-dimensional case, GP-BO is faster than our method BTAO, and one reason is that, when running one configuration in linear SVM is not so costly, it is more computational efficient using GP-BO to fit one GP model than using BTAO to fit three GP models.
We then apply our method BTAO and the other five methods to optimize the hyperparameters of SVM on MNIST \cite{BOHB, Klein2017}. We use a RBF kernel in SVM containing two hyperparameters: the regularization hyperparameter $C$ and the kernel hyperparameter $\gamma$, with $C \in [2^{-10}, 2^{10}], \gamma \in [2^{-10}, 2^{10}]$. For MNIST, we use 50,000 images as the training set, 10,000 images as the validation set, and set the iterations for LT and HT runs to be $(E_l,E_h)=(50,500)$. As Figure \ref{fig:SVM} shows, all the 6 methods have almost the same comparative final performances. The model-based methods TPE, GP-BO, SMAC, BOHB and BTAO reach the best performance much faster than Random Search. Comparing the time to reach the best performance, there is no much difference between BOHB and BTAO. While BTAO is faster in reducing the validation error during the early training period, one possible reason is that BTAO can take the advantage of NLHD with more evenly distributed initialization and judiciously selecting the subsequent configuations on the two-level runs. As for the two fully model-based approaches GP-BO and BTAO, BTAO is two times faster than GP-BO.

\subsection{Fully Connected Networks on MNIST}

\begin{figure}[H]
	\centering
	\includegraphics[width=3.6in]{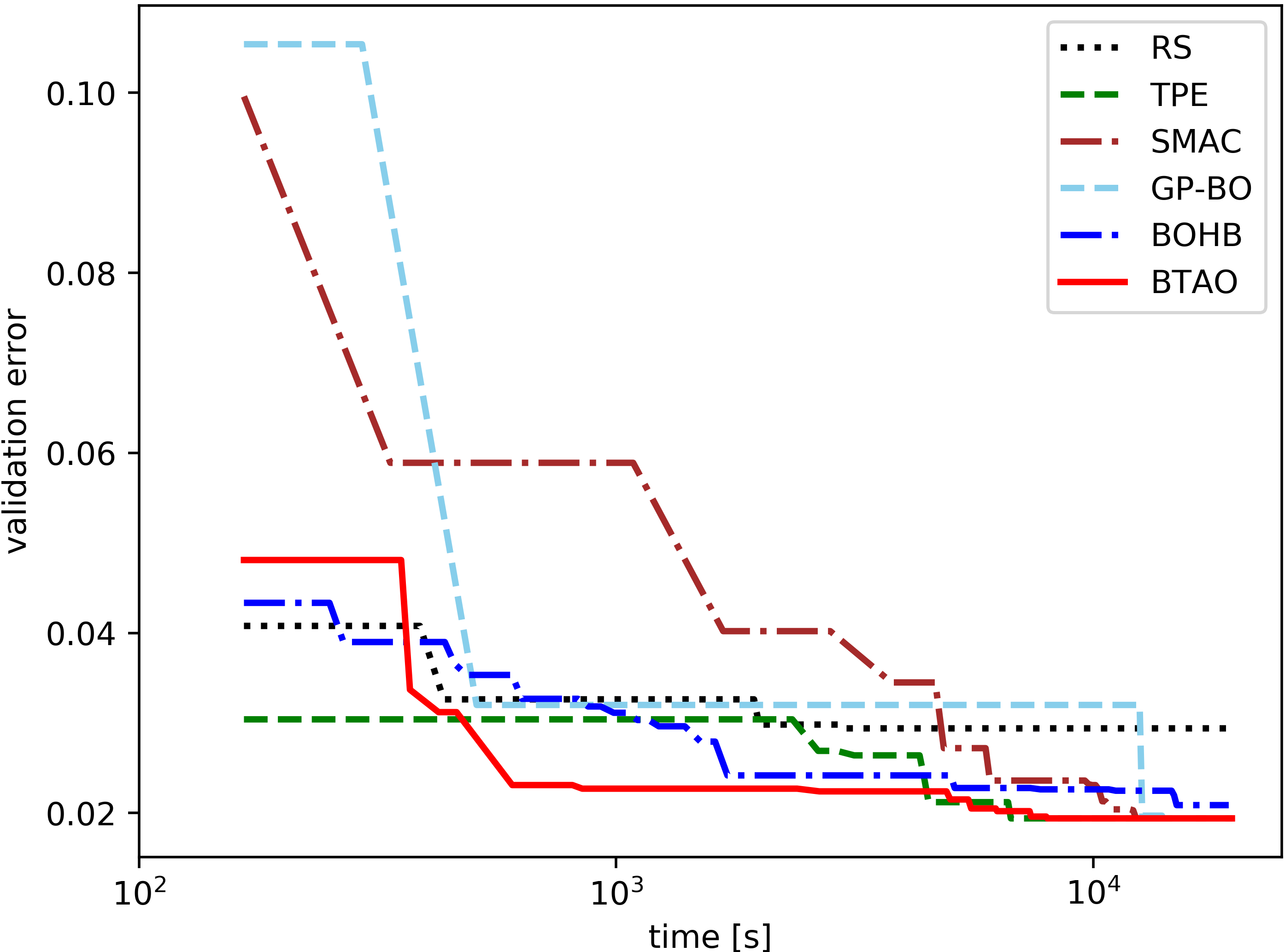}
	\caption{Hyperparameter optimization of a one-layer fully connected network on MNIST.}\label{fig:FCN}
\end{figure}

For this experiment, we optimize three hyperparameters of a one-layer fully connected network on MNIST. The splitting of training and validation set is the same as in the previous experiment. The hyperparameters considered in this experiment are the number of hidden units, batch size and initial learning rate, with their ranges shown in Table \ref{tab:hyper_FCN}. The parameters that define the LT runs and HT runs are $(E_l, s_l, e_l)=(10, 3, 0.001)$ and $(E_h, s_h, e_l)=(50,3,0)$, respectively. In Figure \ref{fig:FCN}, all methods performs much better than Random Search. BTAO, TPE, GP-BO and SMAC achieve almost the same best performance with the best validation error around 0.016, but BTAO is consistently faster than GP-BO and SMAC, and five times faster than SMAC in the middle training period.

%In order to further compare the final results of the methods, as shown in Table \ref{fcn_result_tab}, the first column and the second column listed the best hyperparameter configuration and the best validation error that each method had reached during the same amount of training time, respectively. Both SMAC and BTAO achieved the best validation error 0.0168, but the best hyperparameter configuration that constructed the neural network found by BTAO with 442 hidden units was smaller than that found by SMAC with 479 hidden units.

\begin{table}[h]
	\caption{Hyperparameters for the fully connected networks and their ranges.}
	\label{tab:hyper_FCN}
	\centering
	\setlength{\tabcolsep}{8mm}{
		\begin{tabular}{cc}
			\toprule
			Hyperparameter & Range  \\
			\hline
			batch size &  $[2^{3}, 2^{9}]$ \\
			\hline
			number of hidden units  & $[2^4, 2^9]$  \\
			\hline
			initial learning rate  & $ [10^{-6}, 10^{-2}]$ \\
			\bottomrule
	\end{tabular}}
\end{table}

\begin{comment}
\begin{table}
\centering
\caption{The final results of different methods in FCN.}
\label{fcn_result_tab}
\begin{tabular}{ |c|c|c|}
\hline
\multirow{2}*{Method} & Configuration & \multirow{2}*{Validation error} \\% & \multirow{2}*{Time (s)}\\
& (bach size, \# units, leraning rate) & \\%  &  \\
\hline
RS & $ (481, 452, 9.57  \times 10^{-4})$ & 0.0198 \\% &  70.193\\
\hline
TPE	& $(89, 508, 7.25 \times 10^{-4})$ &	0.0177 \\ % & 52.768\\
\hline
GP-BO & $(199, 507, 1.19 \times 10^{-3} )$ & 0.0170 \\ %& 63.932 \\
\hline
SMAC & $(272, 479, 1.92 \times 10^{-3})$	& \textbf{0.0168} \\% & 71.743\\
\hline 
BOHB &	$(10, 415, 3.6 \times 10^{-4} )$ & 0.0199	\\ % & 337.243\\
\hline 
BTAO & $ (154, 442, 1.11\times 10^{-3})$ & \textbf{0.0168} \\ % & 37.268\\
\hline
\end{tabular}
\end{table}
\end{comment}

\subsection{Convolutional Neural Network on CIFAR-10}

\begin{figure}[H]
	\centering
	\includegraphics[width=3.6in]{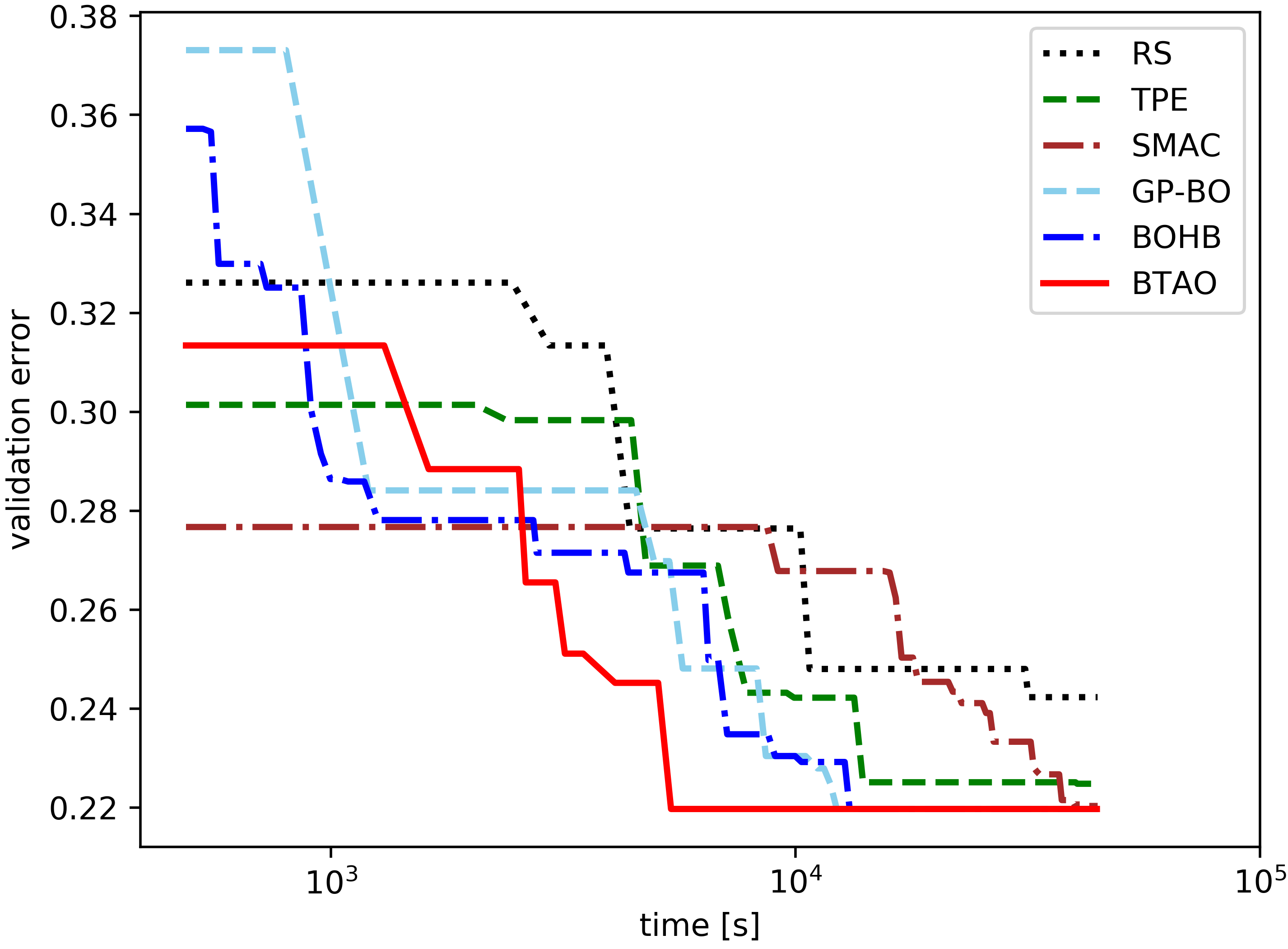}
	\caption{Hyperparameters  optimization of CNN on CIFAR-10.}\label{fig:CNN}
	
\end{figure}

In the last experiment, we evaluate the performance of our method BTAO on a more expensive network and optimize the validation error of a two-layer convolutional neural network (CNN) on CIFAR-10. The hyperparameters of interest for the two-layer CNN are the number of hidden units in the first and second layer, learning rate and dropout rate, with the ranges shown in Table \ref{tab:hyper_CNN}. We split off 10,000 training images from the 50,000 training images as a validation set to evaluate the performance. The LT runs and HT runs are defined by the parameters $(E_l, s_l, e_l)=(10,3,0.001)$ and $(E_h, s_h, e_l)=(50,3,0)$, respectively. The results in Figure \ref{fig:CNN} shows that all model-based methods TPE, SMAC, BOHB, GP-BO and BTAO substantially outperform Random Search, but TPE is not so competitive as the other four model-based methods in terms of the best validation error. BTAO finds the optimal hyperparameters 1.5-1.8 times faster than the other methods.

%According to Table \ref{cnn_result_tab}, SMAC achieved the validation error 0.2174 which was slightly better than BTAO with the validation error 0.2179. For the best hyperparameter configuration of BTAO, the network was smaller and had fewer total number of parameters (weights) than that of SMAC. 

\begin{table}[h]
	\caption{Hyperparameters for CNNs and their ranges.}
	\label{tab:hyper_CNN}
	\centering
	\setlength{\tabcolsep}{8mm}{
		\begin{tabular}{cc}
			\toprule
			Hyperparameter & Range  \\
			\hline
			\# units layer 1  &  $[2^3, 2^{6}]$ \\
			\# units layer 2   & $[2^3, 2^7]$ \\
			initial learning rate  & $ [10^{-6}, 10^{-2}]$ \\
			dropout rate & $[0, 0.5]$ \\
			\bottomrule
	\end{tabular}}
\end{table}

\begin{comment}
\begin{table}
\centering
\caption{The final results of different methods in CNN. Here $h_1, h_2, dr$ and $lr$ represent the number of hidden units in the first layer, the number of hidden units in the second layer, the dropout rate and the learning rate, respectively.}
\label{cnn_result_tab}
\begin{tabular}{ |c|c|c|}
\hline
\multirow{2}*{Method} & Configuration & \multirow{2}*{Validation error} \\ % & \multirow{2}*{Time (s)}\\
& ($h_1, h_2, dr, lr)$ & \\ % &  \\
\hline
RS & $ (60, 107, 0.409, 1.01 \times 10^{-4})$ & 0.2264 \\ % & 315.055\\
\hline
TPE	& $(51, 67, 0.4836, 1.84 \times 10^{-4})$ & 0.2283 \\ %& 238.635\\
\hline
GP-BO & $(63, 127, 0.5, 2.23\times 10^{-4})$	& 0.2186	\\ %&462.588\\
\hline
SMAC & $(63, 103, 0.496, 1.79\times 10^{-4})$ & \textbf{0.2174} \\ %& 626.105\\
\hline 
BOHB &	$(44, 107, 0.496, 1.46 \times 10^{-4} )$ & 0.2195 \\ %& 316.99\\
\hline 
BTAO & $(54, 120, 0.4868, 2.14 \times 10^{-4})$ & \textbf{0.2179} \\ %& 332.568\\
\hline
\end{tabular}
\end{table}
\end{comment}

\section{Conclusion}\label{sec:conclusion}

We introduce BTAO, a fully model-based Bayesian optimization method for hyperparameter optimization. We extend the standard way of modeling the objective function by using a Gaussian process and a truncated Gaussian process to integrate the two-level training performance measurements (i.e., LT and HT runs). We also develop a sequential way to select the promising configurations with robustness and efficiency. Our new method BTAO can both fit the performance profile of the configuration space and find the optimal configurations. Experiments demonstrates the potentials of BTAO in a wide range of ML applications.

\bibliographystyle{plain}
\bibliography{references}

\end{document}